\PassOptionsToPackage{dvipsnames,svgnames, x11names}{xcolor}

\documentclass[10pt, a4paper]{article}

\usepackage[final]{lrec-coling2024} 

\usepackage{soul}
\usepackage{amsmath}
\usepackage{amsthm}
\usepackage{amssymb}
\usepackage{bm}         
\usepackage{tabularx}
\usepackage{multirow}
\usepackage{extarrows}
\usepackage{colortbl}
\usepackage{subfigure}
\usepackage[title]{appendix}
\usepackage{hyperref}

\definecolor{myGreen}{RGB}{161,225,133}

\title{KC-GenRe: A Knowledge-constrained Generative Re-ranking Method Based on Large Language Models for Knowledge Graph Completion}

\name{
	\begin{tabular}{c}
		Yilin Wang$^{1}$, Minghao Hu$^{2,*}$\thanks{* Corresponding author}, Zhen Huang$^{3,*}$, Dongsheng Li$^{3}$, Dong Yang$^{3}$, \\
		Xicheng Lu$^{3}$\\
	\end{tabular}
}

\address{$^{1}$ Defense Innovation Institute, Academy of Military Sciences, \\ $^{2}$ Information Research Center of Military Science, \\ $^{3}$ National Key Laboratory of Parallel and Distributed Computing  \\  
	\{wangyilin14, huangzhen, dsli, yangdong14, xclu\}@nudt.edu.cn, huminghao16@gmail.com\\}

\abstract{
The goal of knowledge graph completion (KGC) is to predict missing facts among entities. 
Previous methods for KGC re-ranking are mostly built on non-generative language models to obtain the probability of each candidate.
Recently, generative large language models (LLMs) have shown outstanding performance on several tasks such as information extraction and dialog systems.
Leveraging them for KGC re-ranking is beneficial for leveraging the extensive pre-trained knowledge and powerful generative capabilities. 
However, it may encounter new problems when accomplishing the task, namely mismatch, misordering and omission. 
To this end, we introduce \textbf{KC-GenRe}, a knowledge-constrained generative re-ranking method based on LLMs for KGC.
To overcome the mismatch issue, we formulate the KGC re-ranking task as a candidate identifier sorting generation problem implemented by generative LLMs.
To tackle the misordering issue, we develop a knowledge-guided interactive training method that enhances the identification and ranking of candidates.
To address the omission issue, we design a knowledge-augmented constrained inference method that enables contextual prompting and controlled generation, so as to obtain valid rankings. 
Experimental results show that KG-GenRe achieves state-of-the-art performance on four datasets, with gains of up to 6.7\% and 7.7\% in the MRR and Hits@1 metric compared to previous methods, and 9.0\% and 11.1\% compared to that without re-ranking.
Extensive analysis demonstrates the effectiveness of components in KG-GenRe.
 \\ \newline  }

\begin{document}

\maketitleabstract

\section{Introduction}
Knowledge graph (KG) stores facts in the form of triples, where each of them is represented as (head entity, relation, tail entity), i.e., $(e_h,r,e_t)$.
However, KGs are generally incomplete as a large number of facts are missing~\cite{kgincomplete}, which hinders the performance of a wide range of applications such as question answering~\cite{saxena2020improving,sun-etal-2022-jointlk}, and recommendation systems~\cite{yang2022KGCLrecommendation}.
Knowledge graph completion (KGC) is therefore a critical task to predict missing facts for improving KG completeness.

Recently, large language models (LLMs), e.g. GPT3~\cite{brown2020language} and LLaMA~\cite{touvron2023llama}, have shown excellent performance on various tasks such as information extraction and dialog system~\cite{wang2023instructuie,touvron2023llama}. 
Meanwhile, several approaches explore the ability of LLMs to perform KGC~\cite{Zhu2023LLMsFK,yao2023exploring}.
However, due to the uncontrollable and diverse nature of the generation  process, reasoning about missing entities directly for query $(e_h,r,?)$ from the unfine-tuned LLMs requires manual assistance to match the output with entities in KG~\cite{Zhu2023LLMsFK}. 
This makes it difficult to automatically obtain answers and perform a comprehensive evaluation on the whole dataset. 
Although there are methods to train LLMs for KGC~\cite{yao2023exploring}, they do not deviate from previous ones~\cite{Chen2022KnowledgeIF,GenKGC-WWW} that are based on small generative language models (LMs), e.g., T5~\cite{Raffel2020t5}. 
Furthermore, prior methods for generative KGC in the domain of commonsense knowledge~\cite{19_COMET,23_KBC} allow to generate new entities not belonging to the given KG, which differs from the conventional KGC task focused here.

Lately, re-ranking using LLMs has received attention on tasks such as information retrieval~\cite{zhu2023large,Pradeep2023RankVicunaZL,ma2023zeroshot}.
Unlike these approaches for re-ranking documents related to a query, we utilize LLMs for re-ranking given candidate entities to implement KGC.
Nevertheless, the task is non-trivial that query is relatively brief, consisting solely of a head entity and a relation. 
Moreover, there is a paucity of knowledge regarding the candidates, which may be limited to their name labels, thereby  increasing the difficulty of the re-ranking process.

\begin{figure}
	\centering
	\includegraphics[width=\linewidth]{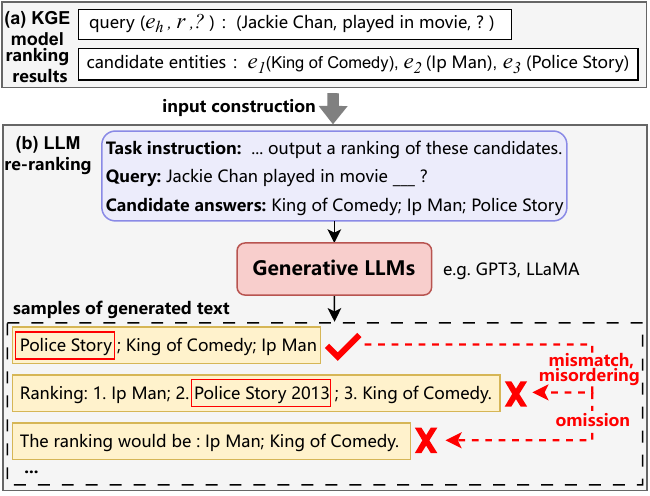}
	\caption{Challenges for KGC re-ranking based on generative LLMs, given query (Jackie Chan, played in movie, ?) and the top-3 candidates, where $e_3$(Police Story) is the target entity.
	}
	\label{fig:example}
\end{figure}

Additionally, to the best of our knowledge, existing re-ranking methods for KGC based on LMs all utilize non-generative models such as BERT~\cite{devlin2018bert} to obtain the probability of each candidate.
Although utilizing generative LLMs for re-ranking can inherit their robust capabilities, such as extensive pre-trained knowledge and flexible task reconstruction, there are new challenges that need to be addressed, as exemplified at the bottom of Figure~\ref{fig:example}:  
(1) \textbf{mismatch}: the generated text contains the candidate entity in KG, but in different textual forms, e.g., generating ``Police Story 2013'' instead of target ``Police Story''; 
(2) \textbf{misordering}: the correct answer is not predicted in the first position;
(3) \textbf{omission}: the output text fails to include all candidates, particularly the target one.
For instance, LLMs may refuse to answer questions by outputting ``I’m sorry, but I don’t have enough information to answer''~\cite{yao2023exploring}, hindering the performance and evaluation of KGC re-ranking.

To tackle the above problems, we propose \textbf{KC-GenRe}, a \textbf{k}nowledge-\textbf{c}onstrained \textbf{gen}erative \textbf{re}-ranking method that fully exploits the potential of generative LLMs to perform KGC re-ranking.
To overcome the mismatch issue, we formalize the re-ranking task as outputting the order of option identifiers corresponding to these candidates, rather than their names.
This eliminates the need for exact text matching of entities and limits the output vocabulary to a known and finite range.
To resolve the misordering issue, we design a knowledge-guided interaction training method that utilizes the inference results from the first-stage knowledge graph embedding (KGE) model to enhance the discernment of candidates and learn their relative order.
To deal with the omission issue, we present a knowledge-augmented constrained inference method that retrieves contextual knowledge to perform generation under control, so as to output legitimate results. 
In summary, our contributions are:
\begin{itemize}
	\item We propose KC-GenRe, a novel knowledge-constrained generative re-ranking model, which is the first to utilize generative LLM for KGC re-ranking as far as we know. 
	\item We design knowledge-guided interactive training and knowledge-augmented constrained inference methods to stimulate the potential of generative LLMs and generate valid ranking of candidates for KGC.
	\item Experiments on four datasets show that KG-GenRe outperforms start-of-arts results, and extensive analysis demonstrates the effectiveness of the proposed components. Datasets and codes have been open sourced at \href{https://github.com/wylResearch/KC-GenRe}{https://github.com/wylResearch/KC-GenRe}.

\end{itemize}

\section{Related Work}

\subsection{Embedding-based KGC}
KGE methods  measure the plausibility of triples by learning low-dimensional embeddings of entities and relations. They are popular in implementing KGC task, which can be broadly divided into three categories: 
distance-based models~\cite{bordes2013translating,TransAt_ijcai2018p596}, 
semantic matching-based models~\cite{trouillon2016complex,balazevic-etal-2019-tucker}, 
and neural network-based models~\cite{schlichtkrull2018modeling,dettmers2018convolutional}.
Auxiliary information such as entity descriptions, hierarchical types 
are often employed to enhance the embedding so that it contains not only structural but also semantic information~\cite{xie2016representation,tkrl2016}.

\subsection{LM-based KGC}
LM-based KGC methods typically utilize the textual form of triples, falling into two categories.
The first type utilizes LMs to perform KGC independently, which can be transformed into a binary classification task for query triple $(e_h,r,e_t)?$~\cite{yao2019kg,kim2020multi}, a matching task to find missing entities for $(e_h,r,?)$~\cite{wang-etal-2022-simkgc}, or a text generation task to output target entities~\cite{kgt5-saxena-etal-2022-sequence,Chen2022KnowledgeIF,GenKGC-WWW,yao2023exploring,Zhu2023LLMsFK}. 
KG-S2S~\cite{Chen2022KnowledgeIF} trains a small LM, specifically  T5~\cite{Raffel2020t5}, to generate a single candidate entity during each inference process for $(e_h,r,?)$.
Although KG-LLM~\cite{yao2023exploring} fine-tunes LLMs like LLaMA~\cite{touvron2023llama} to further achieve triple classification and relation prediction tasks, it solely designs input-output templates.
~\citet{Zhu2023LLMsFK} employ LLMs for zero-shot and few-shot KGC without fine-tuning, but they need manual sampling for answer acquisition and evaluation.

Above methods linearize the triple knowledge into text, resulting in a lack of structural information. 
Hence, the second category combines LMs and KGE models to capture both semantic and structural knowledge. 
They either integrate the LM into KGE methods~\cite{zhang2020pretrain,wang2021structure} or utilize it as an independent re-ranking stage following KGE model~\cite{lovelace2021robust,PKGC-lv-etal-2022-pre}.

\subsection{LM-based KGC Re-ranking}
Existing re-ranking methods are basically built on non-generative LMs to compute the probability of each candidate.
CEAR~\cite{kolluru2021cear} applies BERT~\cite{devlin2018bert} to score a set of candidates together by concatenating their textual forms with query $(e_h,r,?)$.
~\cite{lovelace2021robust} develops a BERT-based student network guided by a first-stage KGE model to score the text consisting of the query and a candidate. 
In addition to the query triple $(e_h,r,e_t)$, PKGC~\cite{PKGC-lv-etal-2022-pre} takes entity definition and attribute as prompts, whose relation templates are manually designed. 
TAGREAL~\cite{jiang2023textTAGREAL} proposes to automatically generate query prompts and retrieve related information from large corpora to construct input.
Different from them, to the best of our knowledge, we are the first to model the KGC re-ranking process utilizing generative LMs, and our contextual knowledge are retrieved from training set.

\subsection{Re-ranking Tasks with LLMs}
Recently, generative LLMs show generalized and superior performance across numerous tasks such as translation, dialogue, and information extraction~\cite{wang2023instructuie,touvron2023llama}.  
It exhibits various capabilities such as zero-shot reasoning, in-context learning, and instruction understanding~\cite{kojima2022large,min2022rethinking,ouyang2022training}.
Meanwhile, in addition to traditional full parameter fine-tuning approach, many techniques have emerged, such as adaptation, prompt learning, and instruction tuning~\cite{hu2021lora,liu2022p,wang2023instructuie}, so as to apply LLMs to various downstream tasks. 
Although there are several methods based on LLMs for re-ranking on information retrieval task~\cite{zhu2023large,Pradeep2023RankVicunaZL,ma2023zeroshot}, as far as we know that we are the first to fine-tune LLMs for re-ranking on KGC task. 
With respect to the challenges faced by KGC, we further propose novel training and inference methods.

\begin{figure*}
	\centering
	\includegraphics[width=\linewidth]{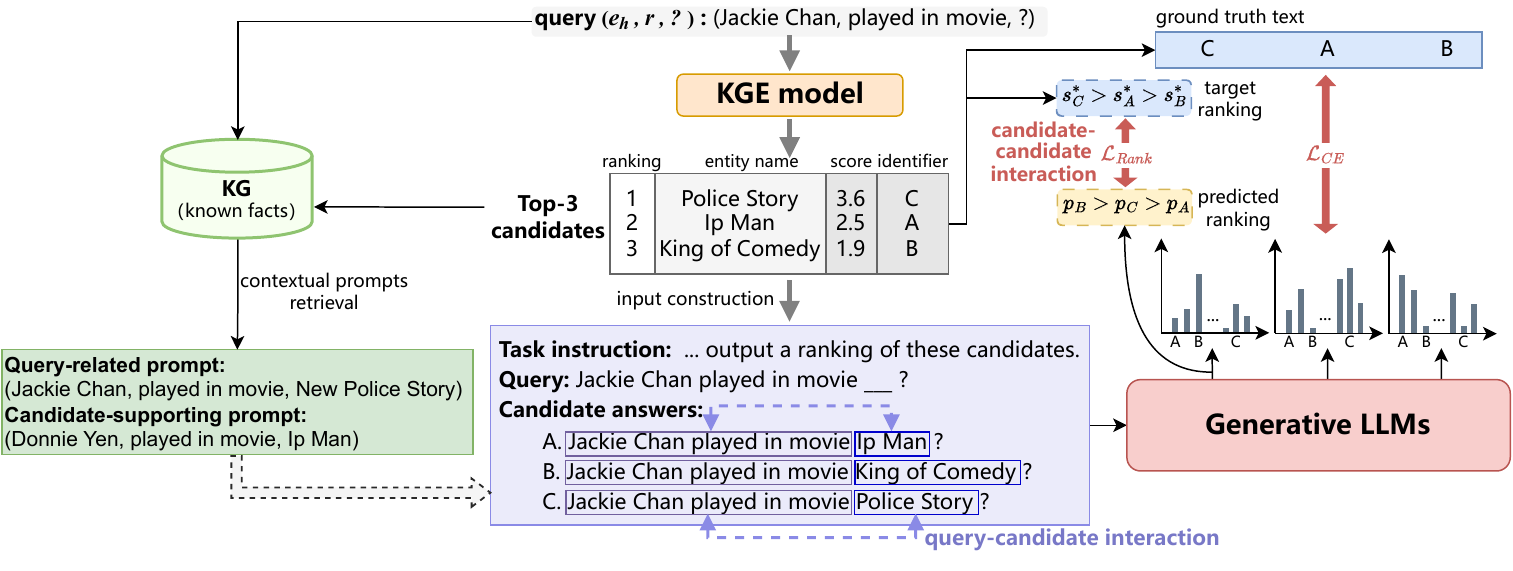}
	\caption{Overview of KC-GenRe, which re-ranks Top-3 candidates predicted by the first-stage KGE model through LLMs for a given query $(e_h,r,?)$.
		Its knowledge-guided interactive training method includes query-candidate interaction and candidate-candidate interaction modules, while its knowledge-augmented constrained inference method includes query-related prompt, candidate-supporting prompt, and constrained option generation modules(omitted in the figure). 
	}
	\label{fig:genre}
\end{figure*}

\section{KGC Re-ranking Formulation}
Let $\mathcal{G} = ( \mathcal{E}, \mathcal{R}, \mathcal{F} )$ be a KG, where $\mathcal{E}$ is the set of entities, $\mathcal{R}$ is the set of relations, and $\mathcal{F}=\{(e_h,r,e_t)|e_h\in \mathcal{E},r\in \mathcal{R},e_t \in \mathcal{E}\}$ is the set of facts. 
Here, $e_h$ and $e_t$ are the head entity and tail entity in a factual triple.
Each entity $e\in\mathcal{E}$ and each relation $r\in\mathcal{R}$ has its own labeled name text, which is a sequence of tokens, denoted as $x^e=(w^e_1, w^e_2,\dots,w^e_{|x_e|})$ and $x^r=(w^r_1, w^r_{2},\dots,w^r_{|x_r|})$, where $|x_e|$ and $|x_r|$ are the numbers of tokens in $x^e$ and $x^r$.
Given a query $(e_h,r,?)$, link prediction task ranks each entity by calculating its score that makes the query hold, so as to achieve KGC.

In the two-stage framework, the first ranking stage typically employs efficient KGE methods to obtain scores for each entity in answering the query $(e_h,r,?)$. 
We denote the top-$K$ predicted candidate entities as $ E_c =\{ e_{t_1}, e_{t_2},\dots, e_{t_K}\}$, and their corresponding scores as $ S_c =\{ s_{1}, s_{2},\dots, s_{K}\}$.
At the second stage of re-ranking, these promising entities in $E_c$ are converted into a sequence along with the query, which serves as the input to KC-GenRe to output their ranking.

\section{KC-GenRe: The Proposed Method}
To harness the capabilities of generative LLMs for KGC, we propose a knowledge-constrained generative re-ranking model, named KC-GenRe, to obtain the ranking of top-$K$ predicted candidate entities for a given query $(e_h,r,?)$, as shown in Figure~\ref{fig:genre}.
In Sec.~\ref{sec:method-reranking-formalization}, we formally introduce the generative KGC re-ranking task proposed here. 
To achieve comprehensive discernment of candidates and learn their relative ranking, we develop the knowledge-guided interactive training method in Sec.~\ref{sec:method-reranking-training}.
For generating effective and legitimate candidate ranking, we design the knowledge-augmented constrained inference method to provide supporting contextual prompt and generation control in Sec.~\ref{sec:method-reranking-inference}.

\subsection{Generative Re-ranking }
\label{sec:method-reranking-formalization}
\paragraph{Input Composition}
Given a query $(e_h,r,?)$ and top-$K$ candidate entities $E_c$ predicted by the first-stage KGE models, we first convert them into input sequence $x_{in}$ by an instruction template $T_{in}$, i.e., $x_{in} = T_{in}(x_q,x_c)$, where $x_q$ and $x_c$ are query sequence and candidate sequence respectively.
We can obtain $x_q$ either by directly concatenating the text of $e_h$ and $r$, or by adopting pre-constructed prompt, e.g., $x_q$ could be ``Jackie Chan played in movie \underline{\hspace{0.5cm}}?'' for query (Jackie Chan, played in movie, ?).
As for candidate sequence $x_c$, it is composed of the given candidate entities $E_c$ with each one equipped with an option identifier selected from the set of option identifiers $O$, such as ``A. Ip Man B. King of Comedy C. Police Story''.

\paragraph{Generation Target}
To address the problem of mismatch, our objective is to generate the order of option identifiers associated with these candidates, instead of their labeled name texts. 
It restricts the output vocabulary to a small and fixed set of candidate identifiers, eliminating text matching with a large number of entities.
The target output text $y$ is a concatenation of identifiers, e.g., ``C A B''.

\subsection{Knowledge-guided Interactive Training}
\label{sec:method-reranking-training}
To address the issue of misordering, we design a knowledge-guided interactive training method, i.e., query-candidate interaction and candidate-candidate interaction, to achieve a thorough identification of candidates and accurately learn their relative ranking by leveraging knowledge from the first-stage KGE models.

\paragraph{Query-Candidate Interaction}
To enhance the discernment of candidates as answers to the query, we explicitly integrate each candidate with the query to form candidate triple in candidate sequence $x_c$, rather than just listing candidate entities. 
Formally, we obtain candidate triple sequence $x_{hrt_i} (i\in\{1,2,\dots,K\})$ for candidate entity $e_{t_i} \in E_c$ by populating its name label to the places where entity is missing in the query sequence $x_q$, e.g., 
$x_{hrt_1}$ = ``Jackie Chan played in movie Ip Man?''.
Hence, the candidate sequence $x_c$ in Sec.~\ref{sec:method-reranking-formalization} would be modified to ``A. Jackie Chan played in movie Ip Man? B. Jackie Chan played in movie King of Comedy? C. Jackie Chan played in movie Police Story?''.

With explicit query-candidate interaction, the model can directly learn the rationality of each candidate triple that represents as a sequence piece in $x_c$. 
In addition, it can establish shorter dependency relationships between the query and each candidate, especially when $K$ is large.

\paragraph{Candidate-Candidate Interaction}
Different from question answering task that only requires identifying correct answers from given candidates, the goal of this paper is to output the sorted result of all candidates.
To learn their relative ranking, we propose to augment their interaction by utilizing the knowledge from first stage, i.e., candidate scores $S_c$, for learning a ranking loss between the target and the predicted sorts.  
First, we employ min-max scaling to normalize $S_c =\{ s_{1},\dots, s_{K}\}$ to the range [0,1] as labeled candidate probabilities $S^* = \{s^*_1, \dots, s^*_K\}$, where $s^*_i$ is the target probability of $e_{t_i}$.
Next, to get the probability predicted by LLMs like LLaMA~\cite{touvron2023llama}, we identify the position of the first option identifier in the generated sequence, from which the model's logits for all option identifiers are taken out as the logits of corresponding candidate entities.
Then, min-max scaling is also applied to the logits to produce predicted probabilities $P = \{p_{1}, \dots, p_{K}\}$, where $p_{i}$ is the probability of option $o_i\in O$, i.e., the predicted probability of candidate $e_{t_i}$ being the correct answer. 
Finally, the ranking loss $\mathcal{L}_{Rank}$ between the target ranking from first stage and the predicted ranking from LLM is:
\begin{equation}
	\small
	\mathcal{L}_{Rank} = \frac{C}{K^2} \sum_{s^*_{i} < s^*_{j}} \max(0, p_{i} - p_{j}), i,j \in \{1,\dots,K\}
	\label{eq:rankloss}
\end{equation}
where $C=100$, and $\frac{C}{K^2}$ is the scaling term used to make $\mathcal{L}_{Rank}$ insensitive to $K$, since the number of additive terms in $\mathcal{L}_{Rank}$ is $K^2/2$.
Specifically, the scaling term equals 1 when $K$ is set to 10.

\paragraph{Training}
We fine-tune the LLM through the following objective:
\begin{equation}
	\label{eq:loss}
	\mathcal{L} = \mathcal{L}_{CE} + \lambda \mathcal{L}_{Rank}
\end{equation}
where weight $\lambda \in [0,1]$ and  $\mathcal{L}_{CE}$ is the cross-entropy loss typically used in generative LLMs.
To construct training samples consisting of input sequence $x_{in}$ and target sequence $y$, we utilize the trained KGE model of the first stage to infer queries in the training set and get the predicted top-$K$ candidates $E_c$ as well as their corresponding scores $S_c$. 
The candidates in $E_c$ are shuffled to form $x_{in}$, and $y$ is obtained by sorting these candidates based on their scores.
Each training triple $(e_h,r,e_t)$ makes up two samples by separately querying $(e_h,r,?)$ and $(?,r,e_t)$, namely queries that missing the tail and the head.
No negative samples need to be constructed.

\subsection{Knowledge-augmented Constrained Inference}
\label{sec:method-reranking-inference}
Although the fine-tuned LLMs can to some extent capture the knowledge stored in KG and learn the output format, there exists omission issue where some or all of the candidates are not generated.
The latter case typically arises from a lack of contextual knowledge, leading to a refusal to generate a ranking.
To  enhance the reasoning ability at inference stage, we retrieve two kinds of knowledge as prompts to assist the input side, namely query-related prompt and candidate-supporting prompt.
Moreover, we design a constrained option generation method at the output side, to ensure that the answer exists and is validly ordered.

\paragraph{Query-related Prompt}
We retrieve query-related training triples to provide contextual knowledge for answering a query.
Formally, each training triple $(e_h,r,e_t) \in \mathcal{F}$ is first converted into a triple sequence $x_{hrt}$, which is part of the training text set $\mathcal{X}_{train}$.
This process is the same as obtaining the candidate triple sequence $x_{hrt_i}$ described in Sec. ~\ref{sec:method-reranking-training}.
Similar to text retrieval methods, we embed the query sequence $x_q$ and all training triples' texts $\mathcal{X}_{train}$ into semantic representation space using an off-the-shelf sentence embedding model, e.g., SBERT~\cite{reimers2019sentence}.
Then we calculate their cosine similarity and use only the top-$K_q$ triple sequences in $\mathcal{X}_{train}$ similar to the given query $x_q$, which are concatenated into a query-related prompt $x^k_q$. 
Denote this process as $F$, which can be formulated as:
\begin{equation}
	\label{eq:x^k_q}
	x^k_q=F(x_q, \mathcal{X}_{train}, K_q)
\end{equation}

\paragraph{Candidate-supporting Prompt}
Taking a step further, we retrieve for each candidate the evidence supporting it as the answer from known training triples.
To achieve this, we take the candidate triple sequence $x_{hrt_i}$ obtained in \emph{query-candidate interaction} as the query and retrieve its top-$K_c$ similar triple sequences from $\mathcal{X}_{train}$ as support for candidate $e_{t_i}$, denoted as $x^k_{e_{t_i}}$, which is analogous to gaining query-related prompt:
\begin{equation}
	\label{eq:x_cand_ti}
	x^k_{{t_i}}=F(x_{hrt_i}, \mathcal{X}_{train}, K_c)
\end{equation}
Then we concatenate the supporting sequences of each candidate to form the candidate-supporting prompt $x^k_{c}$:
\begin{equation}
	x^k_{c}=[x^k_{{t_1}}, x^k_{{t_2}}, \dots, x^k_{{t_K}}]
\end{equation}
where [] means concatenation of texts.
To control the length of $x^k_{c}$ and retain only the most useful supporting information, we additionally set a similarity threshold $\theta \in [0,1]$ during retrieval.
Therefore, Equation~(\ref{eq:x_cand_ti}) can be modified as:
\begin{equation}
	\label{eq:x_cand_ti_new}
	x^k_{{t_i}}=F(x_{hrt_i}, \mathcal{X}_{train}, K_c, \theta)
\end{equation}
Note that $x^k_c$ could be empty text if none of the candidates has supporting information that satisfies the condition. %

\paragraph{Constrained Option Generation}
During decoding, the LLM may suffer from the problem of being unable to output a complete ranking of option identifiers, even with the process of fine-tuning or the provision of context. 
Since the output format in this paper is simplified without generating labeled name texts of candidate entities, we propose to restrict the model to output all option identifiers without duplication, thus yielding a valid ranking. 
Due to the enormous permutation number of option identifiers, instead of using prefix constraints by enumerating all legal outputs~\cite{Chen2022KnowledgeIF}, we directly narrow down the legitimate words from the entire vocabulary to the set of candidate option identifiers that have not appeared, at each position where an option identifier should be generated.

\paragraph{Inference}
Due to the model's instruction understanding capability, we employ a new instruction template $T_{in}^k$ during inference to transform  $x_{in}$ into knowledge-augmented input sequence $x^k_{in}$, which contains query-related prompt $x^k_q$ and candidate-supporting prompt $x^k_c$, noted as $x^k_{in} =T_{in}^k(x_q,x_c,x^k_q,x^k_c)$.
Note that we do not use these two retrieved prompts during the training phase, as they may contain noise that could affect the learning process.
The ordering of each option identifier in the generated text is used as the ranking of the corresponding candidate entity for evaluation.

\begin{table}
	\scriptsize
	\centering
	\setlength\tabcolsep{5.2pt} 
	\renewcommand{\arraystretch}{1.2}
	\begin{tabular}{lrrrrr}
		\hline	
		Dataset & $|\mathcal{E}|$ & $|\mathcal{R}|$ & \# Train & \# Valid & \# Test \\
		\hline
		Wiki27K 	& 27,122 	& 62		& 74,793 	& 10,121	& 10,122\\
		FB15K-237-N & 13,104	& 93 		& 87,282	& 7,041		& 8,226\\
		ReVerb20K	& 11,065	& 11,058	& 15,499	& 1,550		& 2,325\\
		ReVerb45K	& 27,008	& 21,623	& 35,970	& 3,598		& 5,395 \\
		\hline
	\end{tabular}
	\caption{\label{tab:Dataset} 
		Dataset statistics.
	}
\end{table}
\begin{table}
	\scriptsize
	\centering
	\setlength\tabcolsep{5pt} 
	\renewcommand{\arraystretch}{1.2}
	\begin{tabular}{lcccc}
		\hline	
		Dataset  & Wiki27K & FB15K-237-N  & ReVerb20K & ReVerb45K \\
		\hline
		$K$		& 	20	& 20  & 30 	& 30 \\
		$\lambda$ 	&  0.1	& 0.1 & 0.3 & 1.0 \\
		\hline
	\end{tabular}
	\caption{\label{tab:parameters} 
		Values of Hyperparameters. 
	}
\end{table}

\section{Experimental Setup}

\subsection{Dataset}
Following PKGC~\cite{PKGC-lv-etal-2022-pre}, we apply two curated KG datasets named Wiki27K and FB15K-237-N, which are sampled from Wikidata and Freebase. 
We also use two open KG datasets following CaRe~\cite{gupta2019care}, namely ReVerb20K and ReVerb45K, which are extracted from text corpus by open information extraction approach ReVerb~\cite{fader2011identifying}. 
Note that triples in open KG are in the form of \textbf{(head noun phrase, relation phrase, tail noun phrase)}, where the noun phrase (NP) and relation phrase (RP) are not canonicalized.
This means that there exists NPs that link to the same entity, such as ``Microsoft'' and ``Microsoft Corporation'', and RPs that refer to the same relation, e.g., ``be a close friend of'' and ``become good friend with''.
Gold canonical clusters of NPs extracted through Freebase entity linking information~\cite{gabrilovich2013facc1} are provided for evaluating missing tail NPs.
The statistics of these datasets are listed in Table~\ref{tab:Dataset}.
For more details, we refer readers to the related papers.

\begin{table}[t]
	\small
	\setlength\tabcolsep{3pt} 
	\renewcommand{\arraystretch}{1.2}
	\centering
	{
		\begin{tabularx}{\linewidth}{lX}
			\hline	
			$T_{in}$	& Below is an instruction that describes a task, paired with a question and corresponding candidate answers. The questions and candidate answers have been combined into candidate corresponding statements. Combine what you know, output a ranking of these candidate answers.$\setminus$n$\setminus$n
			\#\#\# Question: \sethlcolor{LightSkyBlue}\hl{\{$x_q$\}}$\setminus$n$\setminus$n \sethlcolor{pink}\hl{\{$x_c$\}} 
			\#\#\# Response:  \\
			
			\hline
			$T^k_{in}$		& Below is an instruction that describes a task, paired with a question and corresponding candidate answers. The questions and candidate answers have been combined into candidate corresponding statements. Knowledge related to some candidates will be provided that may be useful for ranking. Combine what you know and the following knowledge, output a ranking of these candidate answers.$\setminus$n$\setminus$n 
			\#\#\# Supporting information: \sethlcolor{yellow}\hl{\{$x^k_q$\}}$\setminus$n$\setminus$n 
			\#\#\# Candidate supporting knowledge: \sethlcolor{myGreen}\hl{\{$x^k_c$\}}$\setminus$n$\setminus$n 
			\#\#\# Question: \sethlcolor{LightSkyBlue}\hl{\{$x_q$\}}$\setminus$n$\setminus$n \sethlcolor{pink}\hl{\{$x_c$\}} 
			\#\#\# Response:  \\
			\hline
		\end{tabularx}
	}
	\caption{\label{tab:template in} 
		Instruction templates of KC-GenRe, where   
		\protect \sethlcolor{LightSkyBlue}\hl{$x_q$}, \protect\sethlcolor{pink}\hl{$x_c$}, \protect\sethlcolor{yellow}\hl{$x^k_q$} and \protect\sethlcolor{myGreen}\hl{$x^k_c$} represent \protect\sethlcolor{LightSkyBlue}\hl{query sequence}, \protect\sethlcolor{pink}\hl{candidate sequence}, \protect\sethlcolor{yellow}\hl{query-related prompt}, and \protect\sethlcolor{myGreen}\hl{candidate-supporting prompt} respectively.
	}
\end{table}

\begin{table*}[!h]
	\scriptsize		
	\centering
	\setlength\tabcolsep{9.5pt} 
	\renewcommand{\arraystretch}{1.2}
	\begin{tabular}{l|cccc|cccc}
		\hline
		\multirow{2}{*}{Model}
		& \multicolumn{4}{c|}{Wiki27K}  &\multicolumn{4}{c}{FB15K-237-N}\\
		& MRR   & Hits@1& Hits@3& Hits@10   & MRR   & Hits@1& Hits@3& Hits@10 \\ 
		\hline
		TransE$^\dagger$~\cite{bordes2013translating} 				& 0.155 & 0.032 & 0.228 & 0.378 	& 0.255 & 0.152 & 0.301 & 0.459 \\
		TransC$^\dagger$~\cite{lv-etal-2018-differentiating}			& 0.175 & 0.124 & 0.215 & 0.339    	& 0.233 & 0.129 & 0.298 & 0.395 \\
		ConvE$^\dagger$~\cite{dettmers2018convolutional} 				& 0.226 & 0.164 & 0.244 & 0.354 	& 0.273 & 0.192 & 0.305 & 0.429 \\
		WWV$^\dagger$~\cite{WWV_ijcai2019}							& 0.198 & 0.157 & 0.237 & 0.365  	& 0.269 & 0.137 & 0.287 & 0.443 \\
		
		TuckER~\cite{balazevic-etal-2019-tucker}			& 0.249 & 0.185 & 0.269 & 0.385 	& 0.309 & 0.227 & 0.340 & 0.474 \\

		RotatE$^\dagger$~\cite{sun2019rotate}							& 0.216 & 0.123 & 0.256 & 0.394 	& 0.279 & 0.177 & 0.320 & 0.481 \\
		\hline
		KG-BERT$^\dagger$~\cite{yao2019kg} 							& 0.192 & 0.119 & 0.219 & 0.352  	& 0.203 & 0.139 & 0.201 & 0.403  \\
		
		LP-RP-RR$^\dagger$~\cite{kim2020multi}						& 0.217 & 0.138 & 0.235 & 0.379  	& 0.248 & 0.155 & 0.256 & 0.436 \\
		
		PKGC$^\dagger$~\cite{PKGC-lv-etal-2022-pre} 					& \underline{0.285} & \underline{0.230} & \underline{0.305} & \textbf{0.409}  	& \underline{0.332} & \underline{0.261} & \underline{0.346} & \underline{0.487} \\	
		
		\hline
		
		KC-GenRe  	& \textbf{0.317} & \textbf{0.274} & \textbf{0.330} & \underline{0.408} 		& \textbf{0.399} & \textbf{0.338} & \textbf{0.427} & \textbf{0.505} \\

		\hline
	\end{tabular}
	\caption{\label{tab:main results-ckg} 
		Link prediction results on two curated KGs. 
		Best results are in bold and second best are underlined.  
		[$^\dagger$]: results are taken from PKGC~\cite{PKGC-lv-etal-2022-pre}.
	}
\end{table*}
%
\begin{table*}[!h]
	\scriptsize		
	\centering
	\setlength\tabcolsep{4.2pt} 
	\renewcommand{\arraystretch}{1.2}
	\begin{tabular}{l|ccccc|ccccc}
		\hline
		\multirow{2}{*}{Model}
		& \multicolumn{5}{c|}{ReVerb20K}  &\multicolumn{5}{c}{ReVerb45K}\\
		&	MRR & MR     & Hits@1& Hits@3& Hits@10      & MRR   & MR     & Hits@1& Hits@3& Hits@10 \\ 
		\hline
		TransE~\cite{bordes2013translating} 				& 0.138 & 1150.5 & 0.034 & 0.201 & 0.316 		& 0.202 & 1889.5 & 0.122 & 0.243 & 0.346 \\
		
		ComplEx~\cite{trouillon2016complex} 				& 0.038 & 4486.5 & 0.017 & 0.043 & 0.071        & 0.068 & 5659.8 & 0.054 & 0.071 & 0.093 \\
		
		R-GCN~\cite{schlichtkrull2018modeling} 	& 0.122 & 1204.3 &   -   &   -   & 0.187        & 0.042 & 2866.8 &   -   &   -   & 0.046  \\
		
		ConvE~\cite{dettmers2018convolutional} 				& 0.262 & 1483.7 & 0.203 & 0.287 & 0.371 		& 0.218 & 3306.8 & 0.166 & 0.243 & 0.314 \\
		
		KG-BERT~\cite{yao2019kg} 							& 0.047 & {420.4}  & 0.014 & 0.039 & 0.105        & 0.123 & 1325.8 & 0.070 & 0.131 & 0.223\\
		
		RotatE~\cite{sun2019rotate}							& 0.065 & 2861.5 & 0.043 & 0.069 & 0.108 		& 0.141 & 3033.4 & 0.110 & 0.147 & 0.196 \\
		
		
		PairRE~\cite{chao2020pairre} 						& 0.213 & 1366.2 & 0.166 & 0.229 & 0.296 		& 0.205 & 2608.4 & 0.153 & 0.228 & 0.302 \\
		
		ResNet~\cite{lovelace2021robust}					& 0.224 & 2258.4 & 0.188 & 0.240 & 0.292 		& 0.181	& 3928.9 & 0.150 & 0.196 & 0.242  \\
		BertResNet-ReRank~\cite{lovelace2021robust} 		& 0.272 & 1245.6 & 0.225 & 0.294 & 0.347  		& 0.208 & 2773.4 & 0.166 & 0.227 & 0.281  \\

		\hline
		CaRe~\cite{gupta2019care} 				& 0.318 &  973.2 & -     &  -    &  0.439       & 0.324 & 1308.0 &  -    &  -    & 0.456 \\
		OKGIT~\cite{talukdar2021okgit} \ 			& 0.359 & 527.1  & 0.282 & 0.394 &  0.499       & 0.332 & \underline{773.9} & 0.261 & 0.363 & 0.464\\	
		OKGSE~\cite{OKGSE22} 						& {0.372} & 487.3 & {0.291} & {0.408} & \underline{0.524} & {0.342} & \textbf{771.1} & {0.274} & {0.371} & {0.473} \\

		CEKFA~\cite{CEKFA_ijcai23} 				& \underline{0.387} & \underline{416.7} & \underline{0.310} & \underline{0.427} & 0.515 & \underline{0.369} & 884.5  & \underline{0.294} & \underline{0.409} & \underline{0.502} \\ 
		\hline
		
		KC-GenRe  & \textbf{0.408} & \textbf{410.8} & \textbf{0.331} & \textbf{0.450} & \textbf{0.547} & \textbf{0.404} & 874.1 & \textbf{0.332} & \textbf{0.444} & \textbf{0.534} \\

		\hline
	\end{tabular}
	\caption{\label{tab:main results} 
		Link prediction results on two open KGs. 
		Best results are in bold and second best  are underlined. 
	}
\end{table*}

\subsection{Evaluation Metrics}
\label{sec:exp-metric}
We verify our approach on the link prediction task under filtered setting~\cite{bordes2013translating} by standard ranking metrics: mean rank (MR), mean reciprocal rank (MRR), and hits at $n$ (Hits@$n$), $n=\{1,3,10\}$.
Note that in Open KG, we follow CaRe~\cite{gupta2019care} to evaluate the rank of canonical clusters for the target NP.

\subsection{Experimental Settings}
At the first ranking stage, for Wiki27K and FB15K-237-N, we apply TuckER~\cite{balazevic-etal-2019-tucker} as the KGE model and set the embedding dimension $d=256$ following PKGC~\cite{PKGC-lv-etal-2022-pre}.
For ReVerb20K and ReVerb45K, we follow CEKFA~\cite{CEKFA_ijcai23} and employ its KGE model, noted as CEKFA-KFARe, where $d=768$.

During the second stage, we freeze the parameters of the KGE model and perform inference on the training set to obtain training samples for KC-GenRe.
The instruction templates used in experiments are listed in Table~\ref{tab:template in}.
Please note that we do not consider designing alternative instructions, as this is not the focus of this paper.
We use manually constructed relational prompts following PKGC~\cite{PKGC-lv-etal-2022-pre} to convert triples into natural sentences on Wiki27K and FB15K-237-N datasets, so as to obtain $x_q$, $x_{hrt_i}$ and $x_{hrt}$.
While on ReVerb20K and ReVerb45K, we directly concatenate NPs and RPs in a triple into sentence, given that they are expressed in natural language.

Built on LLaMA-7b~\cite{touvron2023llama}, KC-GenRe is fine-tuned by QLORA~\cite{dettmers2023qlora} approach. 
The re-ranking number $K$ is chosen from \{10,20,30\}, and its value remains the same during both training and testing, unless specifically stated.
Loss weight $\lambda$ is picked from \{0, 0.1, 0.3, 0.5, 0.7, 0.9, 1.0\}. 
The best settings of hyperparameters are listed in Table~\ref{tab:parameters}.
KC-GenRe is fine-tuned for a maximum of 3 epochs, with a batch size of 16 and a learning rate of $1e$-$4$ on all datasets.
Same as PKGC~\cite{PKGC-lv-etal-2022-pre}, we adopt entity definition as auxiliary knowledge during training on Wiki27K and FB15K-237-N.
As a result, during inference, we do not additionally use query-related prompt and candidate-supporting prompt as contextual knowledge on these two datasets.
While on ReVerb20K and ReVerb45K, the query-related prompt and candidate-supporting prompt are both utilized to assist in the inference process, and the values of $K_q$, $K_c$, $\theta$ are empirically set to 3, 3, 0.8 without tuning.
We employ the pre-trained all-mpnet-base-v2 SBERT model~\cite{reimers2019sentence} to encode texts into sentence embeddings for retrieval in Eq.(\ref{eq:x^k_q}) and Eq.(\ref{eq:x_cand_ti_new}).
Greedy search decoding strategy is applied.
All experiments are conducted in Pytorch and on one 80G A800 GPU.

\subsection{Baselines} 

For curated KG datasets Wiki27K and FB15K-237-N, we adopt several models for comparison, including  
(1) KGE-based methods:
TransE~\cite{bordes2013translating}, 
ConvE~\cite{dettmers2018convolutional},  
TuckER~\cite{balazevic-etal-2019-tucker}, 
RotatE~\cite{sun2019rotate}, 
TransC~\cite{lv-etal-2018-differentiating}, 
WWV~\cite{WWV_ijcai2019}, 		
where the last two use concept and definition information, respectively.
(2) LM-based methods: KG-BERT~\cite{yao2019kg}, LP-RP-RR~\cite{kim2020multi}, and PKGC~\cite{PKGC-lv-etal-2022-pre}, where the last one uses entity definition information.

For open KG datasets ReVerb20K and ReVerb45K, the comparison include: 
(1) methods in curated KG: 
TransE~\cite{bordes2013translating},  
ComplEx~\cite{trouillon2016complex}, 
R-GCN~\cite{schlichtkrull2018modeling}, 
ConvE~\cite{dettmers2018convolutional}, 
KG-BERT~\cite{yao2019kg},
RotatE~\cite{sun2019rotate},
PairRE~\cite{chao2020pairre}, 
the two-stage method~\cite{lovelace2021robust}, notated as BertResNet-ReRank, and the query encoding module in it, notated as ResNet;
(2) methods in Open KG: CaRe~\cite{gupta2019care}, OKGIT~\cite{talukdar2021okgit}, OKGSE~\cite{OKGSE22} and CEKFA~\cite{CEKFA_ijcai23}.
Among them, KG-BERT, OKGIT, OKGSE, BertResNet-ReRank, and CEKFA are LM-based methods, with the latter two employing a two-stage reranking architecture.

\section{Experimental Results}

\subsection{Main Results}
\label{sec:results-main}
By targeting the re-ranking of candidate option identifiers and performing constrained option generation, KC-GenRe thoroughly resolves the issues of mismatch and omission in generative KGC based on LLMs. 
While the misordering problem can be assessed through the metrics presented in Table~\ref{tab:main results-ckg} and Table~\ref{tab:main results}.
It can be seen that KC-GenRe outperforms existing works  on both curated and open KGs. 
In Table~\ref{tab:main results-ckg}, we gain absolute improvements of 3.2\% and 6.7\% for MRR, 4.4\% and 7.7\% for Hits@1 on Wiki27K and FB15K-237-N compared to PKGC, which owns the same KGE ranking model (TuckER) as ours.
Compared to TuckER, our re-ranking method KC-GenRe obatins increases of 6.8\% and 9.0\% for MRR, 8.9\% and 11.1\% for Hits@1.
In Table~\ref{tab:main results}, KC-GenRe achieves higher performance than previous methods with improvements of  2.1\% and 3.5\% for MRR, 2.1\% and 3.8\% for Hits@1 on ReVerb20K and ReVerb45K, respectively.  
The suboptimal MR on the ReVerb45K may stem from a few poor predictions (e.g., predicted rankings exceeding 10), thus enlarging the mean number of rankings. 
Nevertheless, KC-GenRe achieves superior performance in terms of MRR and Hits@1, which are considered more reliable and accurate.

From the tables, we can learn that the LM-only approach may not be comparable to traditional KGE approaches.
For example, the Hits@1 metrics implemented by KG-BERT~\cite{yao2019kg} are lower than Tucker~\cite{balazevic-etal-2019-tucker} by 6.4\% and 8.9\% on Wiki27K and FB15K-237-N, and are lower than ConvE~\cite{dettmers2018convolutional} by 18.9\% and 9.6\% on ReVerb20K and ReVerb45K, respectively.
By combining KGE and LM to perform coarse ranking and fine-grained re-ranking respectively, it is possible to ensure high efficiency and recall in the ranking stage, and to obtain further accurate predictions through LMs.

\subsection{Ablation Study}
\label{sec:results-ablation}
Table~\ref{tab:ablation study} and  Table~\ref{tab:ablation study-ckg} show impacts of each component proposed in KC-GenRe, namely query-candidate interaction (QCI), candidate-candidate interaction (CCI), query-related prompt (QP), candidate-supporting prompt (CP), and constrained option generation (CG). 
The first field (Base) in these tables represents the baseline KGE model without re-ranking, and the last two fields are re-ranking models based on generative LLMs.
Specifically, entry ``1'' indicates the re-ranking method that directly fine-tunes the LLMs without using all proposed components.

It can be found that removing individual component or their combinations can lead to performance decreases.  
And the most significant improvements can be observed when all components are adopted, with gains of 4.1\%, 5.2\%, 6.8\% and 9.0\% in MRR on ReVerb20K, ReVerb45K, Wiki27K and FB15K-237-N respectively, compared to the KGE model without re-ranking (Base). And the Hits@1 metric achieves the most significant improvement on FB15K-237-N by 11.1\%. 

Concretely, (1) Directly fine-tuning the generative LLMs for re-ranking without using any components of KC-GenRe (entry 1) can still be effective on Wiki27K and FB15K-237-N. 
However, it brings no performance improvement or even a drop of 1.7\% and 1.8\% in MRR on ReVerb20K and ReVerb45K, compared to the KGE model (Base).
This illustrates the necessity of exploring methods based on LLM that are applicable to KGC task.
(2) The query-candidate interaction is clearly critical for KC-GenRe, with an increased MRR of 2.5\% and 3.9\% on ReVerb20K and ReVerb45K respectively, compared to simply listing all candidates (entry 2). %
It indicates that combining each candidate with the query can fully learn the possibility of each candidate completing the query.
(3) The ranking loss calculated by the candidate-candidate interaction (entry 3) is also helpful.
For instance, on FB15K-237-N, it makes increases in MRR and Hits@1 metrics of 3.8\% and 5.4\% respectively, illustrating the usefulness of learning relative order between candidates.
(4) In Table~\ref{tab:ablation study}, by utilizing both query-related prompt and candidate-supporting prompt together (entry 6) in inference, KC-GenRe achieves 3.8\% and 4.4\% MRR boosts.
This implies that the retrieved relevant training triples indeed provide useful contextual knowledge so that the reasoning ability of LLM can be fully exploited.
When removing one of these two prompts (entry 4 and 5), the performance of KC-GenRe decreases slightly, indicating that they may contain overlapping extracted triplets.
(5) In Table~\ref{tab:ablation study-ckg}, the removal of entity definition prompt (entry 4) has a notable effect on both datasets, emphasizing the importance of providing contextual knowledge for reasoning.
(6) It is worth noting that the performance drop is pronounced when only constrained option generation is removed (entry 7 and entry 5 in two tables), even may be slightly worse than entry ``Base''. 
We analyze the experimental results and find that it is better to use constrained option generation together with candidate-candidate interaction.
This could be due to the fact that if we utilize the logits of option identifiers to calculate the candidate-candidate interaction loss and influence the probability learning of each option identifier during training, a corresponding decoding using the logits of these option identifiers (i.e., constrained option generation) is required during inference.
Failure to do so may disrupt the original decoding balance of LLMs and result in decreased performance.
In conclusion, these results highlight the effectiveness of KC-GenRe in enhancing KGC re-ranking based on generative LLMs through the utilization of the proposed components.

\begin{table}[t] 
	\scriptsize		
	\centering
	\setlength\tabcolsep{2.2pt} 
	\renewcommand{\arraystretch}{1.1}
	\begin{tabular}{lccccc|cc|cc}
		\hline
		\multirow{2}{*}{} & \multirow{2}{*}{QCI}& \multirow{2}{*}{CCI}
		& \multirow{2}{*}{QP} & \multirow{2}{*}{CP} & \multirow{2}{*}{CG}
		& \multicolumn{2}{c|}{ReVerb20K}  & \multicolumn{2}{c}{ReVerb45K}\\
		& & & & & & MRR & Hits@1   & MRR & Hits@1 \\ 
		\hline
		Base & & & & & & 0.367 & 0.288 & 0.352 & 0.274 \\
		\hline
		1 & & & & & &  0.350 & 0.263 & 0.334 & 0.246 \\

		2 & & $\checkmark$ & $\checkmark$& $\checkmark$ & $\checkmark$ & 0.383 & 0.304 & 0.365 & 0.282 \\

		3  & $\checkmark$ & & $\checkmark$ & $\checkmark$ & $\checkmark$ &  0.405 & 0.326 & 0.381 & 0.298 \\ 
		4 & $\checkmark$ & $\checkmark$ & & $\checkmark$ & $\checkmark$ &  0.399 & 0.320 & 0.397 & 0.323 \\    

		5 & $\checkmark$ & $\checkmark$ & $\checkmark$ & & $\checkmark$ &  0.403 & 0.328 & 0.400 & 0.327 \\    

		6 & $\checkmark$ &$\checkmark$ & & & $\checkmark$ &  0.370 & 0.284 & 0.360 & 0.271 \\  

		7 & $\checkmark$ & $\checkmark$ &  $\checkmark$ & $\checkmark$ & &  0.367 & 0.289  & 0.352 & 0.273 \\   

		\hline
		KC-GenRe & $\checkmark$ &  $\checkmark$ &  $\checkmark$ &  $\checkmark$ &  $\checkmark$  & 0.408 & 0.331 &  0.404 & 0.332 \\  
		
		\hline
	\end{tabular}
	\caption{\label{tab:ablation study} 
		Ablation results on open KGs, where ``Base'' is CEKFA-KFARe~\cite{CEKFA_ijcai23}.}
\end{table}
\begin{table}[t] 
	\scriptsize		
	\centering
	\setlength\tabcolsep{2.6pt} 
	\renewcommand{\arraystretch}{1.2}
	\begin{tabular}{lcccc|cc|cc}
		\hline
		\multirow{2}{*}{Model} & \multirow{2}{*}{QCI}& \multirow{2}{*}{CCI}& \multirow{2}{*}{DP} & \multirow{2}{*}{CG}
		& \multicolumn{2}{c|}{Wiki27K}  & \multicolumn{2}{c}{FB15K-237-N}\\
		& & & & & 	MRR & Hits@1     & MRR & Hits@1 \\ 
		\hline
		Base & & & & & 0.249 & 0.185 & 0.309 & 0.227  \\
		\hline
		1& & & & & 0.283 & 0.227 & 0.329 & 0.248 \\ 
		
		2& & $\checkmark$ & $\checkmark$ &  $\checkmark$ & 0.314 & 0.268 & 0.340 & 0.257 \\

		3& $\checkmark$ & & $\checkmark$ &  $\checkmark$ & 0.311 & 0.266 &  0.361 & 0.284\\ 

		4& $\checkmark$ & $\checkmark$ & & $\checkmark$ & 0.298 & 0.249  & 0.353 & 0.277  \\

		5& $\checkmark$ & $\checkmark$ & $\checkmark$ &  & 0.245 & 0.178  &  0.303 & 0.218 \\    

		\hline
		KC-GenRe  & $\checkmark$ &  $\checkmark$ &  $\checkmark$ &   $\checkmark$ & 0.317  & 0.274 & 0.399  &  0.338\\  

		\hline
	\end{tabular}
	\caption{\label{tab:ablation study-ckg} 
		Ablation results on curated KGs, where ``Base'' denotes TuckER~\cite{balazevic-etal-2019-tucker} and ``DP'' represents entity definition prompt.}
\end{table}

\subsection{Influences of Re-ranking Number}
\label{sec:results-k value}
To investigate the effects of the re-ranking number $K$, we conduct experiments in two cases: 
(1) the number of candidates for re-ranking during inference is larger than that during training;
(2) the number of candidates for re-ranking during inference is the same as that during training. 
For case (1), we train the model with top-10 ($K$=10) candidates and re-rank top-$K$ at inference, while for case (2), we train and test the model with top-$K$ candidates.

\begin{figure}[t]
	\setlength{\abovecaptionskip}{0.cm}
	\subfigure[ReVerb20K] 
	{
		\includegraphics[width=3.65cm]{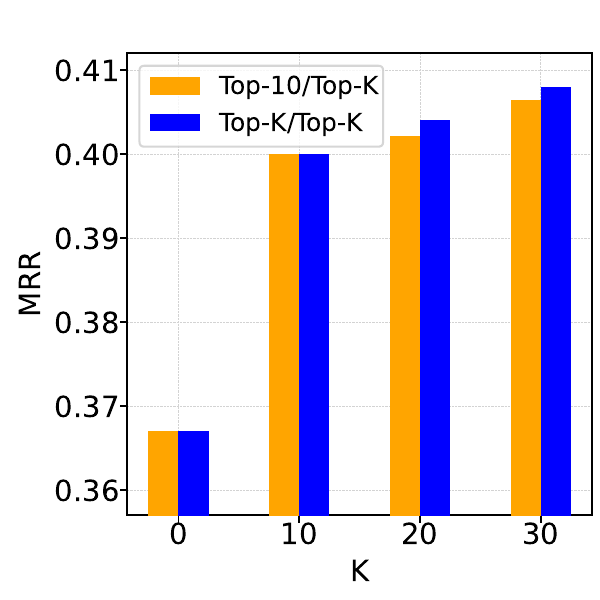}
		\label{fig:topk-20k}
	}\hspace{-2mm}
	\subfigure[ReVerb45K]
	{
		\includegraphics[width=3.65cm]{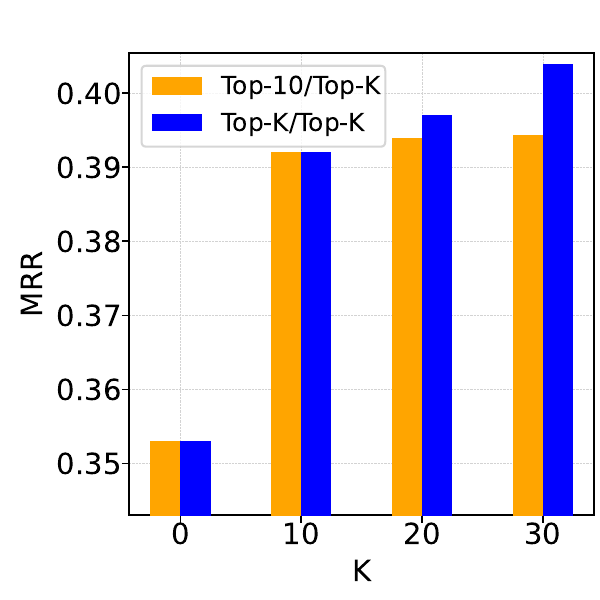}
		\label{fig:topk-45k}
	}	
	\caption{Effects of re-ranking number $K$. The front and back of the slash in the legend represent the values of $K$ during training and testing, respectively.
	}
	\label{fig:freq distribution}
\end{figure}

Figure~\ref{fig:freq distribution} shows the results for case (1) (orange) and case (2) (blue). 
It is obvious that increasing the re-ranking number in both cases leads to performance improvements on both datasets.
This is mainly due to the increase of samples containing correct answers in candidates.
In addition, the results of orange bars indicate that the model trained to rank the top-$K$ candidates has the ability to rank beyond the top-$K$.
However, in Figure~\ref{fig:topk-45k}, they rise slightly with $K$ increased and clearly show poorer performance than blue bars.
It is reasonable because training on more candidates, i.e., negative answers, allows for the learning of more knowledge and the achievement of higher performance.
This is also the reason why the results of case (2) is better than that of case (1).

\subsection{Effects of Candidate-candidate Interaction}
\label{sec:results-rankloss}
\begin{figure}[t]
	\setlength{\abovecaptionskip}{0.cm}
	\subfigure[ReVerb20K] 
	{
		\includegraphics[width=3.65cm]{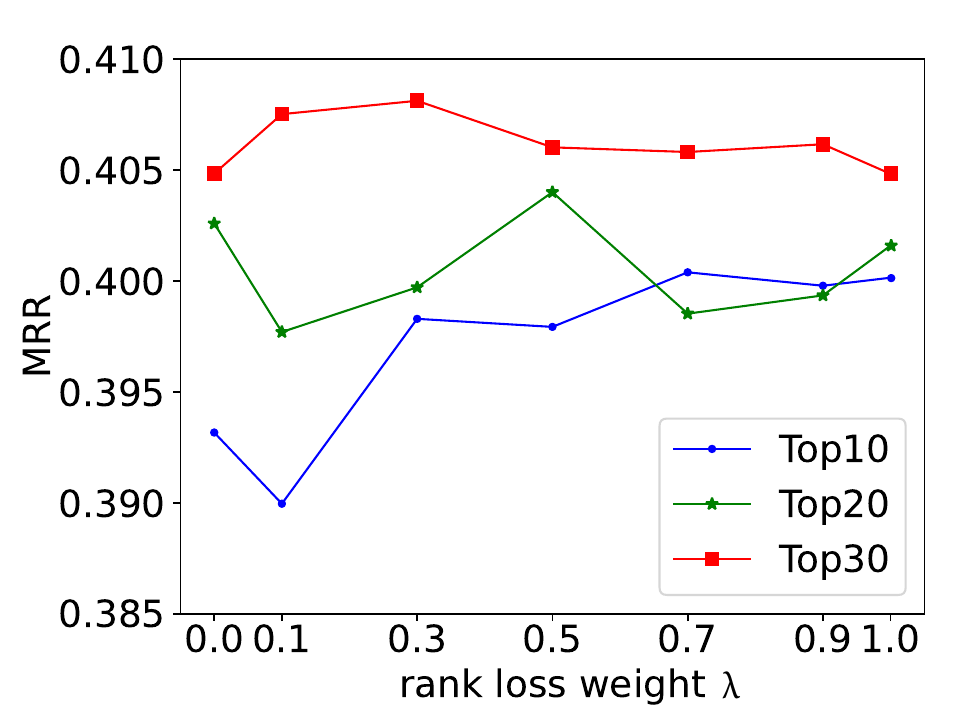}
		\label{fig:rankloss-20k}
	}\hspace{-2mm}
	\subfigure[ReVerb45K]
	{
		\includegraphics[width=3.65cm]{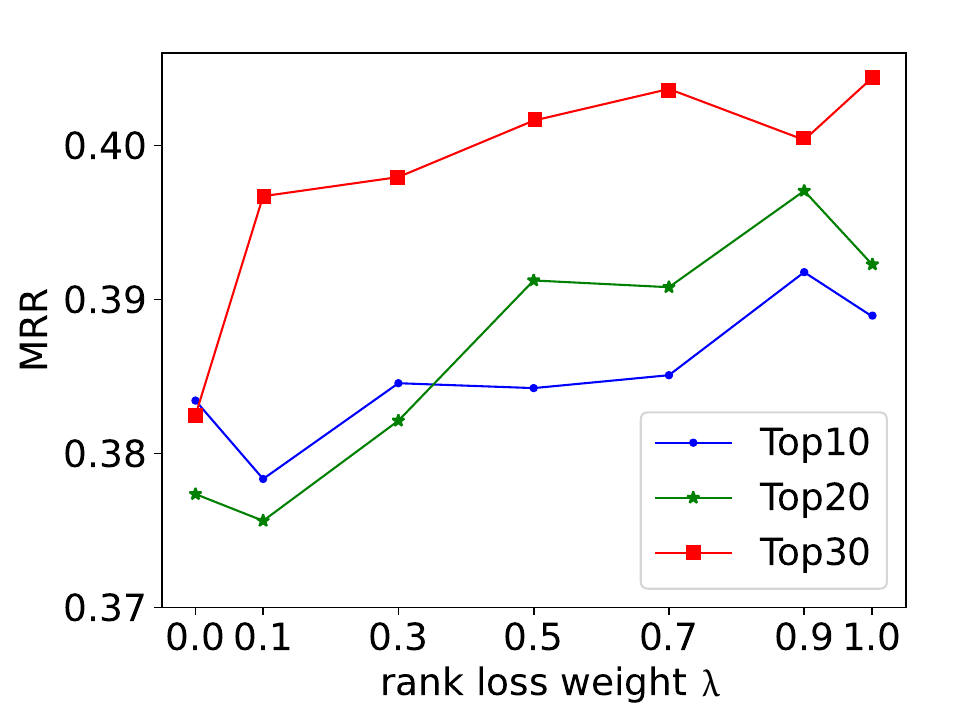}
		\label{fig:rankloss-45k}
	}	
	\caption{Influences of weight $\lambda$ in Eq.(\ref{eq:loss}) with different re-ranking number $K$ (Top-$K$).
	}
	\label{fig:rankloss}
\end{figure}

We compare the influences of weight $\lambda$ for ranking loss $\mathcal{L}_{Rank}$ in candidate-candidate interaction when training and re-ranking with different top-$K$.
As shown in Figure~\ref{fig:rankloss}, the prediction performance is able to be improved after applying candidate-candidate interaction. 
Although the distributions on these datasets are not the same, we can learn that the overall performance is generally better as $K$ gets larger.
Besides, we find that performance drops when $\lambda$ is small ($\lambda=0.1$) with $K\leq20$, while it boosts with $K > 20$. 
This may be due to the fact that sorting difficulty rises with the increase of re-ranking number $K$, leading to a larger ranking loss $\mathcal{L}_{Rank}$, so that a smaller value of $\lambda$ could balance it with $\mathcal{L}_{CE}$ and get commendable performance.

\subsection{Impacts with Different LLMs}
\label{sec:results-large lm}
We compare the results of KC-GenRe using different LLMs, as shown in Table~\ref{tab:llm}. 
Various LLMs exhibit varying levels of performance, e.g., LLaMA2-7b~\cite{touvron2023llama2} achieves higher Hits@1 than LLaMA-7b on ReVerb20K and ReVerb45K by 1.0\% and 0.4\%, respectively.
Increasing the scale of LLMs may bring gains, such as LLaMA-65b gets a 1.7\% Hits@1 increase on ReVerb45K over LLaMA-7b.
However, the lift may be slight.
This could be due to the limited or unmemorized increase of knowledge directly related to the query in the corpus used for pre-training the LLM, resulting in no remarkable improvement in the ability to answer factual questions.
Additionally, a 7B model may be sufficient to achieve optimized results in KGC task~\cite{2023KGCLLM}.

\begin{table}[t]
	\small		
	\centering
	\setlength\tabcolsep{7pt} 
	\begin{tabular}{l|cc|cc}
		\hline
		\multirow{2}{*}{LLM}
		& \multicolumn{2}{c|}{ReVerb20K}  &\multicolumn{2}{c}{ReVerb45K}\\
		&				MRR &  Hits@1  & MRR   &  Hits@1 \\ 
		\hline
		LLaMA-7b 	& 0.400 & 0.324 & 0.392 & 0.325 \\
		LLaMA2-7b 	& 0.406 & 0.334 & 0.397 & 0.329 \\
		LLaMA-13b 	& 0.403 & 0.331 & 0.392 & 0.322 \\
		LLaMA-65b 	& 0.400 & 0.326 & 0.404 & 0.342 \\
		\hline
	\end{tabular}
	\caption{\label{tab:llm} 
		Link prediction results of KC-GenRe implemented with different LLMs when $K=10$.}
\end{table}

\section{Conclusion}
This paper introduces KC-GenRe, a knowledge-constrained generative re-ranking model for KGC.
To tackle mismatch, misordering, and omission issue, we formulates the task as a candidate identifier sorting generation problem and design a knowledge-guided interactive training method 
as well as a knowledge-augmented constrained inference method.
KC-GenRe can enhance the identification and relative ranking of candidates, and generate valid results with supporting prompts. 
Experimental results demonstrate its superior performance and highlighting its effectiveness for KGC.

\section{Acknowledgements}
This work has been partly supported by the National Natural Science Foundation of China
(Grant No. 62025208 and Grant No. 62376284), and the Xiangjiang Laboratory Foundation (Grant No. 22XJ01012).

\section{Bibliographical References}

\bibliographystyle{lrec-coling2024-natbib}
\bibliography{main}



\end{document}